# What do We Learn by Semantic Scene Understanding for Remote Sensing imagery in CNN framework?

Haifeng Li, Jian Peng, Chao Tao*, Jie Chen, Min Deng

School of Geosciences and Info Physics, Central South University

Email: lihaifeng@csu.edu.cn, kingtaochao@csu.edu.cn*(Corresponding Author)

*Abstract* -Recently, deep convolutional neural network (DCNN) achieved increasingly remarkable success and rapidly developed in the field of natural image recognition. Compared with the natural image, the scale of remote sensing image is larger and the scene and the object it represents are more macroscopic. This study inquires whether remote sensing scene and natural scene recognitions differ and raises the following questions: What are the key factors in remote sensing scene recognition? Is the DCNN recognition mechanism centered on object recognition still applicable to the scenarios of remote sensing scene understanding? We performed several experiments to explore the influence of the DCNN structure and the scale of remote sensing scene understanding from the perspective of scene complexity. Our experiment shows that understanding a complex scene depends on an in-depth network and multiple-scale perception. Using a visualization method, we qualitatively and quantitatively analyze the recognition mechanism in a complex remote sensing scene and demonstrate the importance of multi-objective joint semantic support.

## 1. Introduction

Scene is the meaningful combination of multiple objects, environments, and contexts. Compared with object recognition, scene understanding not only needs to identify the targets, it should also understand the distribution of targets in a scene. Thus, the recognition pattern is complex. For example, we recognize a cat in the natural image classification task, which only requires describing the cat's characteristics (texture, color, and shape). Caring about what exists around the cat is unnecessary; however, when adapted in the bedroom, we may have to pay close attention to beds, wardrobes, and other common objects and their spatial interrelationships.

Remote sensing scene has wider imaging range, farther imaging distance, and larger feature dimension (such as mountains, rivers, etc.), and it is susceptible to resolution and atmospheric refraction and other factors. The fine-grained information of the object, which leads to many differences in contrast the natural scene, is difficult to express, and the natural scene likely has a smaller scale and abundant characteristics. For example, the categories in Place 2 (Zhou, Garcia et al. 2014) are mainly indoor scenes (bathroom and office), and these scenes contain a mass of features in rich details. A specific office chair model can also be observed, which is impossible in remote sensing scene. However, understanding remote sensing scenes is never simpler than natural scenes(Desachy 1995, Mattar and Al-Rewihi 1998). Understanding remote sensing scenes is usually plagued by multiple scenes sharing the same objects. For example, there are roads and buildings in business and residential areas, and the differences are often slightly reflected in their density and architectural style. This confusion on spatial distribution pattern makes scene recognition more challenging. Therefore, analyzing the transfer between natural image and remote sensing scene is necessary in understanding the remote sensing scene, and this understanding guides the design of the identification method that captures the specialty of itself.

As the basic task of scene understanding, scene recognition plays an important role in the application of computer vision, and related research has greatly progressed recently(Dixit, Chen et al. 2015, Yin, Jiao et al. 2015, Zhong, Cui et al. 2015, Rangel, Cazorla et al. 2016). Traditional scene recognition methods mainly use the manually designed algorithm to extract the feature to construct the bag of words (BOW) model for scene identification; (Lazebnik, Schmid et al. 2006)proposed spatial pyramid matching (SPM) combined with BOW for scene recognition, and (Quattoni and Torralba 2009) modeled the spatial structure of a scene for indoor scene recognition. These methods strongly depend on the extraction of local features and complex coding strategies. Inspired by human visual mechanism, deep convolutional neural network (DCNN) could achieve automatic hierarchical feature extraction, and it has significantly succeeded in image recognition. (Shen, Lin et al. 2016) explored the problem of information attenuation in DCNN depth from the perspective of information theory and proposed the relay network. (Wang, Guo et al. 2016) considered the diversity between scene and object recognition and introduced a priori knowledge to DCNN model. These successes indicate that DCNN can accomplish natural scene-centered identification tasks.

However, does the current network still works well in the tasks of remote sensing scene recognition? Although previous works (Langkvist, Kiselev et al. 2016, Sevo and Avramovic 2016) have proven that DCNN can be applied to remote sensing scenarios well, its actual satisfaction of our needs remains distant. To make a model work better, we must seize the commonality between remote sensing and natural scenes, wherein rich semantic and scale information on remote sensing scene exist, while considering the difference between them.

Human visual system understands scene by observing it, obtaining information and then abstracting knowledge. In some scenes (such as parks), we need longer observed time and more attention to re-organize information and then extract knowledge; however, for some scenes (such as forest), understanding is completed in an instant. This finding implies that investing on the human visual understanding processes of different scenes is not equitable. To judge which scenes need more visual input to understand them, we propose the concept of scene complexity, which estimates how difficult a scene can be discerned by the human visual system. Limited by the plight that the current dataset is rarely defined from the complexity of the scene, we utilize crowdsourcing to build a mesoscale dataset and make it the basic dataset in the follow-up work.

Is there a mechanism similar to the human visual system in DCNN? Intuitively, the depth of the network greatly affects model performance (Liu, Shen et al. 2014, Kounalakis and Tsampikos 2015, Song, Dai et al. 2016). Depth means that the information transfer in more levels and longer distance, which corresponds to the visual attention time; the size of receptive field (Gal, Hamori et al. 2004, Coates and Ng 2012) determines the DCNN extracts feature in which scale; the different sizes of the receptive field actually corresponds to how far human observes a scene. Note that complex scene understanding often requires observing multiple fields, whether there exists a mechanism in DCNN that multi-objects jointly support the scene understanding. Using a visualized approach, we analyze it qualitatively and quantitatively.

In this paper, we focus on the task of remote sensing scene recognition, revealing the differences between sensing scene and natural scene recognitions based on DCNN, standing on it to explore the pattern as DCNN identifies the remote sensing scene. The main contributions of this paper are as follows:

- We analyzing how the depth of network and the scale of receptive field influence the performance of remote sensing scene understanding tasks, and reveal that using the same fixed depth and scale CNN network for all the types of scene understanding resulting in limited performance. Based on these findings, we suggest that addressing this scale bias from different complexity scene is critical to improve remote sensing scene understanding performance。This finding also inspires us designing a scale-specific network architectures may be a better way for remote sensing image understanding.

- We demonstrate the importance of joint multi-objective semantic support for fine-grained remote sensing scene understanding by analyzing the response of class activation maps in CNNs. This finds shows that Scene and object recognition are two closed related visual tasks, and solve these tasks in an integrated fashion may be a better way for remote sensing image understanding.

## 2. Construction of Experimental Dataset

The quality of the training dataset is an important factor facilitating the performance of the model(Xiao, Hays et al. 2010, Yu, Seff et al. 2015). Superior remote sensing scene dataset not only merely refers to large data, it can also include diverse semantic information (Handa, Patraucean et al. 2015). As especially emphasized, we analyze the influence of network depth and the scale of receptive field from the perspective of scene complexity. For this reason, based on existing remote sensing dataset, we select AID (Xia, Hu et al. 2016) and then sort 22 categories of scene, in which complexity is more distinguishable as our basic dataset (Figure 1). The dataset contains 360 samples per class; each sample is a size of 600×600 RGB image, noting that in order to balance the image size and hardware limits, we resize the original image to $256 \times 256$ in the experiment.

On the basis of the remote sensing interpretation knowledge, the dataset is divided into three superclasses: low, moderate, and high complexities. To objectively evaluate the scene complexity of samples, we invited 10 volunteers to select 10 to 15 samples randomly and score 10−1 (Figure 1) according to the level of complexity. In this paper, we defined the score range from 1 to 4 as low complexity, 4 to 7 as moderate complexity, and 8 to 10 as high complexity and then add up the final score to evaluate the grade of the complex scene.

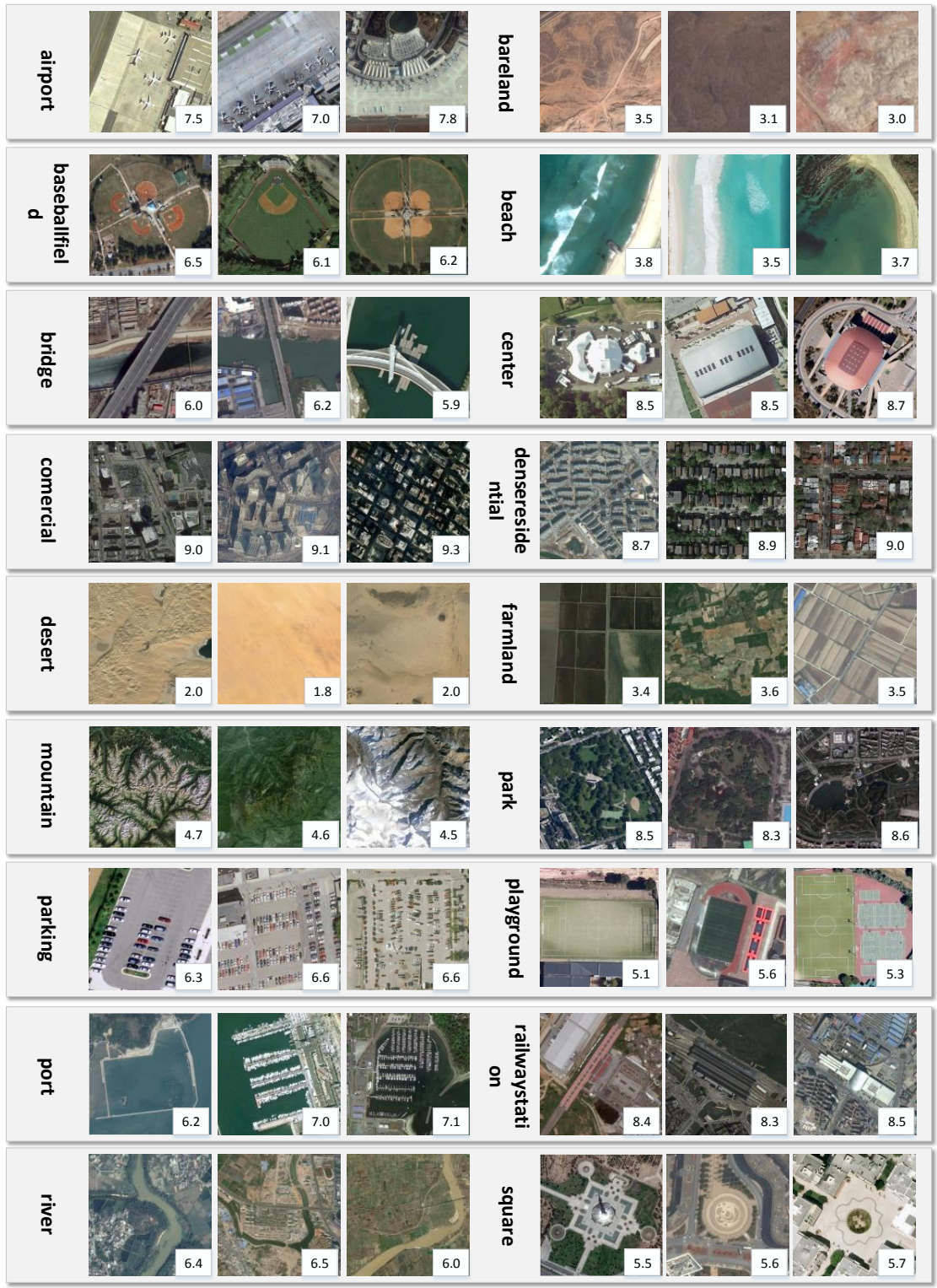

Figure 1. Samples with different scene complexity score

Table 1 provides the scene complexity evaluation results based on subjective visual scoring. I, II, and Ⅲ denote low, middle, and high complexity, respectively.

Table 1. Statistics of different categories in the complexity score

| Categories | Mean | Mode frequency | Guide by knowledge |
|---|---|---|---|
| Bare land | 3.2 | I | I |
| Beach | 3.7 | I | I |
| Desert | 1.9 | I | I |
| Farmland | 3.4 | I | I |
| Forest | 2.7 | I | I |
| Meadow | 1.9 | I | I |
| Baseball field | 6.1 | III | II |
| Bridge | 5.8 | II | II |
| Parking | 6.4 | II | II |
| River | 6.1 | II | II |
| Mountain | 4.6 | II | I |
| Playground | 5.0 | II | II |
| Stadium | 5.2 | II | III |
| Airport | 7.4 | III | III |
| Center | 8.6 | III | III |
| Commercial | 9.3 | III | III |
| Dense residual | 8.9 | III | II |
| Square | 6.5 | III | III |
| Park | 8.2 | III | III |
| Port | 6.4 | III | III |
| Railway station | 8.1 | III | III |
| Viaduct | 7.9 | III | III |

In this study, we use two statistical methods to evaluate the results: the first method takes the mean and then classifies it according to a threshold, whereas the second method classifies it and then takes mode. The results of the two statistical methods are essentially similar in addition to the differences in individual classes (such as playground). The complexity evaluation of major classes is substantially uniform (Figure 2).

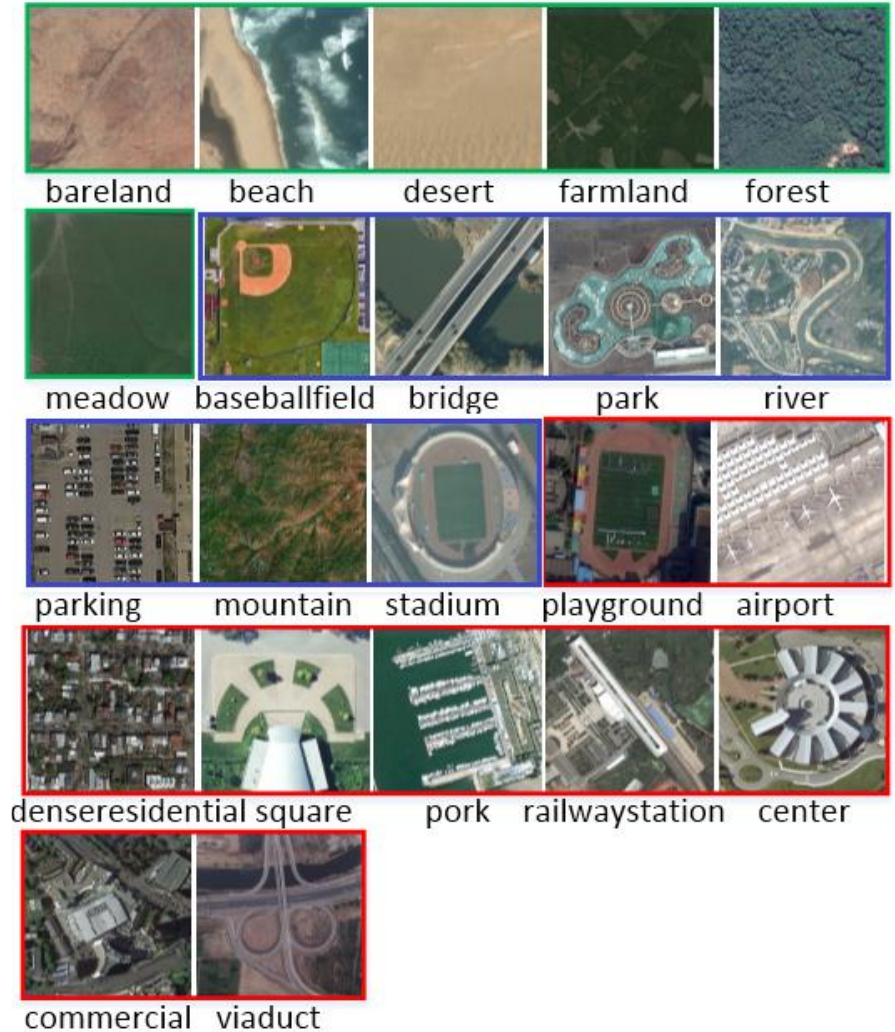

Figure 2. AID-22 scene dataset contains 22 classes of categories; the green, blue, and red rectangles represent the low-complexity, moderately complex, and highly complex scenes

## 3. How Does Network Depth Influence Scene Semantics Understanding?

Abundant research (Krizhevsky, Sutskever et al. 2012, Simonyan and Zisserman 2014) has shown that the deeper the network, the stronger the representation power of the model. However, empirical evidence shows that merely increasing network depth does not effectively improve model performance(He, Zhang et al. 2015, Srivastava, Greff et al. 2015). Intuitively, the needs of the time human eye to observe would change in scenes with different complexities, that is, there are differences in the degree of difficulty when identifying scenes with different complexities. Therefore, we analyze the influence of network depth on the remote sensing scene recognition task from the scene complexity. Experimental results show that different complexity scenarios require various network depths and that highly complex scene recognition will benefit from depth nevertheless, in identifying the low complexity scene; this benefit is not obvious or even harmful. However, deepening the network generally does contribute in improving network performance.

### 3.1 Experiment design in network depth

To analyze the influence of DCNN depth on the task of remote sensing scene recognition, we choose Alex 3/4/5 Conv-Net (Krizhevsky, Sutskever et al. 2012) as our benchmark models. These models are the same except for the use of 3/4/5 convolution layers. To exclude our experimental results for only a specific network, analogously, repeating experiments on VGG 16/19 Net (Simonyan and Zisserman 2014), networks are composed of 13/16 convolution layers and 3 fully connected layers. A unified size of convolutional kernel that eliminates the interference of other possible factors is used.

Experiments use the Caffe (Jia, Yangqing et al. 2014) deep learning framework and train models on the AID-22 Scene dataset. Note that considering the training difficulties caused by deep network, we pre-train and then fine-tune the VGGNet on the ImageNet dataset to obtain the final model.

### 3.2 Experimental analysis and conclusion

We used overall accuracy (OA) to evaluate the model. Table 2 shows the results of testing on VGG-net after 20 w iterations. The table indicates that in the task of remote sensing scene recognition, model performance is affected by various factors and not only network depth. Simply deepening network depth do not improve model performance. Similar to the AlexNet series network (Table 3), 3conv-layers performs best and 5conv-layers performs worst, and simply increasing network depth damages network performance. The VGG performance is remarkable relative to AlexNet mainly due to the depth of the network and the uniform size of the convolution kernel; therefore, the model has a stronger ability to abstract representation and better captures the details of image features.

Table 2. Overall accuracy tested in AID-22 Scene using VGG-net

| Model | Train accuracy (%) | Test accuracy (%) |
|---|---|---|
| VGG-16 | 96.15 | 94.10 |
| VGG-19 | 96.89 | 94.91 |

Table 3. Overall accuracy tested in AID-22 Scene using Alex-net

| Model | Train accuracy | Test accuracy |
|---|---|---|

|               | （%）  | （%）  |
| ------------- | ----- | ----- |
| Alexnet-3conv | 85.32 | 78.02 |
| Alexnet-4conv | 85.67 | 75.47 |
| Alexnet-5conv | 81.58 | 75.07 |

In the task of remote sensing scene recognition, network performs unequally on various scenes, and the performance of different networks is not the same on one-scene recognition. We test VGG-net with different depths to recognize each scene class (Figure 3). Varying model depths perform differently, for example, the VGG-16 has preferable bare land recognition and weak desert scene recognition results, whereas the VGG-19 model works better than the VGG-16 on playground recognition.

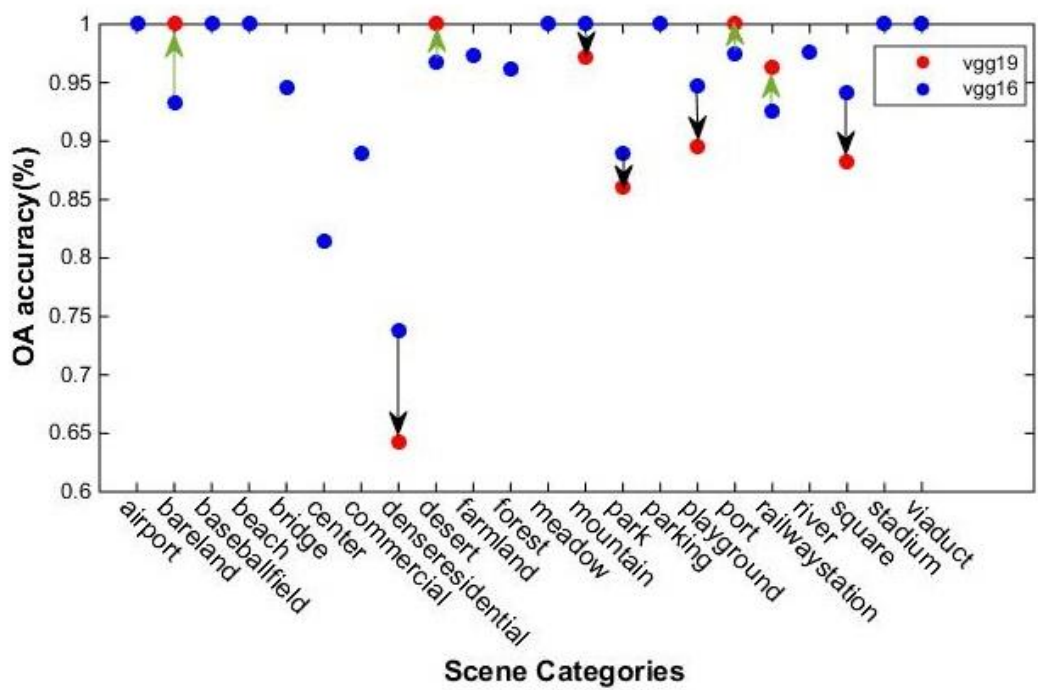

Figure 3. Results of the recognition of different depth VGG networks on various scene categories. The blue dots indicate the result of the VGG-16 testing on the dataset, whereas the green dots indicate the VGG19. The arrow direction shows the change in the accuracy of a certain scene on both models

Various network depths have different recognition abilities on different scenes. To answer whether these scenes have something in common and if model depth and recognition ability are related for these scenes, we further analyze the effect of depth on the recognition of different scenarios from the complexity perspective. Different degrees of complexity of a scene have different sensitivities on network depth, and the change of network depth has minimal effect on the recognition task of simple scenes, but it has great influence on the recognition result of a complex scene. Deepening network depth can effectively improve the ability of the model in recognizing complex scenes, whereas reducing it significantly weakens the ability of the model in identifying complex scenes. Figure 4 shows the overall accuracy of different network depths over the three complexity superclasses. On the AlexNet and VGG networks, a simple scene is better identified in a shallow network, and when deepening the depth, the result worsens; for moderate complex scenes, the overall accuracy increases when the network becomes deeper. Notably, the recognition accuracy of highly complex scenes tested on the VGG network improves as depth increases, but AlexNet begins to decline. This finding intuitively explains that the power of representation in AlexNet is limited on this highly complex scene.

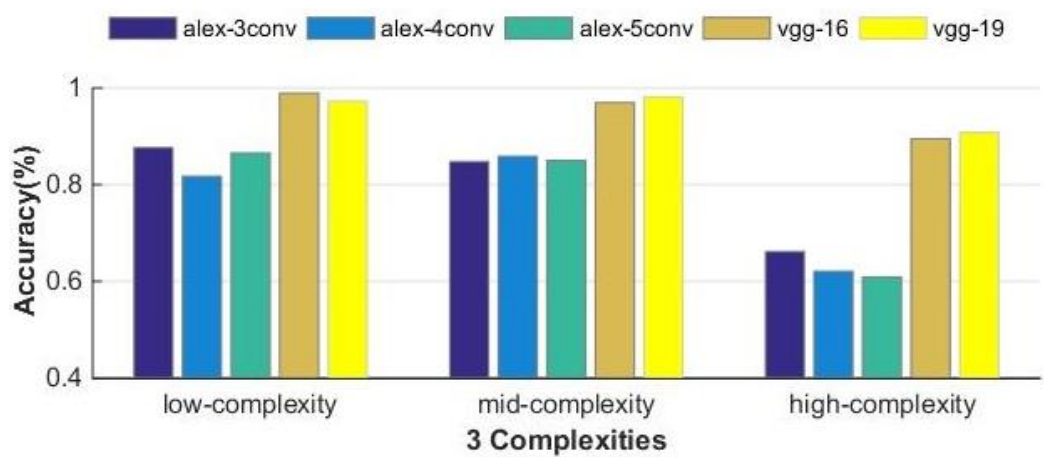

Figure 4. Recognition results of different complexity scenarios for various depth networks. The horizontal axis represents various depth models tested on different complexity remote sensing scene; the vertical axis represents test accuracy

## 4. How Does Network Scale Changes Influence Scene Semantic Understanding?

Remote sensing scenes with large scale and objects in the scene are more ambitious(Wu and Li 2009, Blaschke 2010). Therefore, grasping overall characteristics is significant for semantic understanding of remote sensing scene (Benz, Hofmann et al. 2004, Zhang, Xia et al. 2015). Multiple classes share objects, for example, dense residential and center scene are greatly similar, and they both contain roads and buildings; the difference between them is mainly reflected in the building style (Figure 5). This semantic difference under similar content becomes a challenge in remote sensing scene understanding, in which learning the overall characteristics of the model and the spatial distribution pattern between objects are required; thus, it is necessary to consider this multi-scale property. We analyze the influence of scale factors on DCNN in remote sensing scene recognition. The experimental results show that learning characteristics of multiple scales is salutary in the scene recognition process.

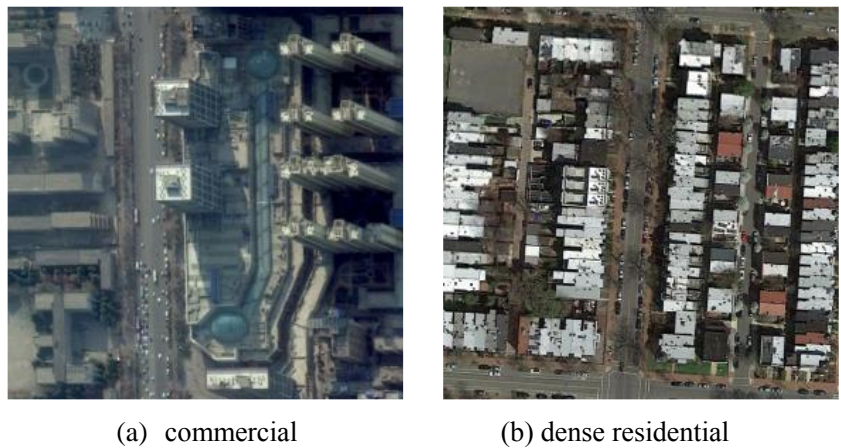

(a) commercial　　　　　　　(b) dense residential

Figure 5. Two similar scenes in content but with different semantic indications

### 4.1 Single- and multi-scale network designs

The scope of a receptive field determines the current layer of the model extract feature in which scale, and it can be calculated by the size of the convolution kernel. The inception structure (Szegedy, Liu et al. 2015) stacks the convolution layers with different kernel sizes into a module and concatenates the features of multiple-scale input into the next module. Given the structural simplicity, we designed a number of single-scale inception networks excluding other possible factors of interference.

We defined the layer input to the inception as layers 1 and 2, which sets uniform convolution kernel sizes of 1×1, 3×3, and 5×5. To ensure that each inception structure learns single-scale feature (Table 4), we use zero padding to create the input and output of the feature map consistent.

Table 4. Single-scale network design in convolution kernel size

| Inception | GoogleNet | Inception (1×1) | Inception (3×3) | Inception (5×5) |
|---|---|---|---|---|
| Previous layer | | | | |
| Layer 1 | 1×1/3×3(p) | 1×1/3×3(p) | 1×1/3×3(p) | 1×1/3×3(p) |
| Layer 2 | 3×3/5×5/ 1×1 | 1×1(3) | 3×3(3) | 5×5(3) |
| Filter concatenation | | | | |

### 4.2 Experimental analysis and conclusion

We utilize OA and kappa coefficient to evaluate the performance of the model because it reflects the error reduction of classification results better than random classification. Experimental results show that the DCNN-based remote sensing scene recognition involves learning multiple scale features. The characteristics of the single scale are insufficient to represent all the semantic information of the scene. Differences in the representation ability of different scale features for the remote sensing scene simultaneously exist, and a larger scale of a feature has stronger expressive ability. Table 5 shows the recognition result of the different scale inceptions embedded in the networks. The kappa value of the GoogleNet model with multiple scale features is higher than that of other single-scale models. The inception 5×5 model with a large convolution kernel scale works best in single-scale models; the smallest scale inception 1×1 model recognition has the worst drops. Thus, it can be deduced that remote sensing scene recognition depends on the characteristics of multiple scales, which mainly adapt to large-scale scenes. Although most information are expressed by coarse-grained features, it still requires some fine-grained features.

Table 5. OA and kappa coefficient tested in multiple-scale inception embedded model

| Model | GoogleNet (多尺度) | Inception $1\times1$ | Inception $3\times n$ | Inception $5\times5$ |
| --- | --- | --- | --- | --- |
| OA | 0.8329 | 0.6863 | 0.7761 | 0.7815 |
| Kappa | 0.8269 | 0.6708 | 0.7651 | 0.7706 |

The identities of different complexity scenes depend on the different scales of feature representation, and complex scene should be represented from multiple scale features. Figure 6 shows the manner in which different inception models identify different complexity scenes; the overall accuracy of multi-scale inception model is higher than other single-scale models for most of the scene recognition. Moreover, the multi-scale inception model outperforms others in highly complex scene recognition tasks, and the accuracy in some of the scenes even increases more than 30%. In analyzing the performance of different scale models on various complexity scenarios (Figure 7), the small-scale model behaves badly on the highly complex scene. As the scale grows, the precision curve rises, noting that the multi-scale model rises fastest and that a similar phenomenon on mid-complexity scene occurs. The accuracy curve of the simple scene peaks in the smaller scale model indicates that the complex remote sensing scene should be represented in a larger scale. The recognition of complex scenes, which requires multiple scales to support, is critical and which simple scenes do not necessarily need.

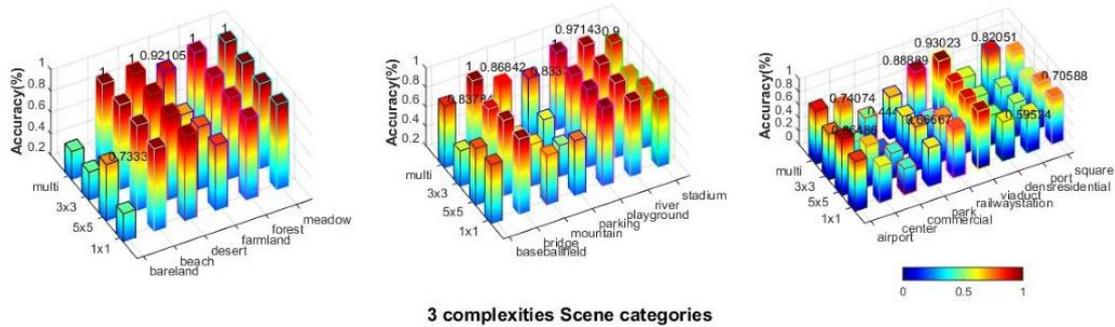

Figure 6. Results of different inception models identify various complex scene classes. The left is the test results of a simple scene, the middle is the moderate complex scene, and the right is the highly complex scene. Warmer colors indicate higher test accuracies; x-axis represents the category of scene, y-axis represents different scales of the model, and z-axis represents the test accuracy

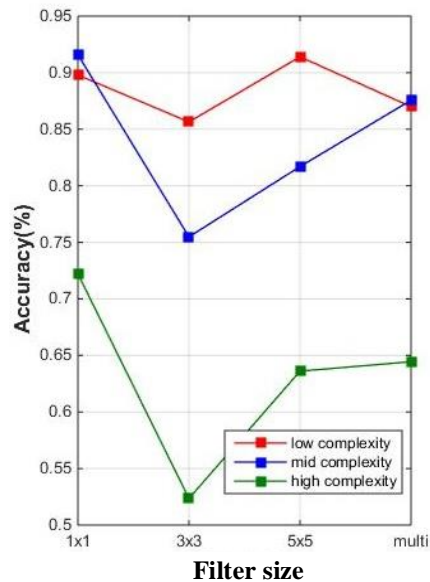

Figure 7. Different inception models recognize different complex scenes. Left: simple scene, mid: moderate complex scene, and right: highly complex scene. Horizontal axis indicates 1×1, 3×3, 5×5, and the original multi-scale inception network, and the vertical axis represents the accuracy of the model tested on a class of scenes

## 5. How Does Combining Target Objects Support Scene Semantic Understanding?

People complete the understanding of a scene by extracting information from different viewpoints in face of complex scenes (Saeed 2015). The current study (Girshick, Donahue et al. 2013, Zhou, Khosla et al. 2014) proves that the model learns a scene by encoding the distribution of objects; however, the feature representation that the model has learned remains unclear. In this feature representation, we use a visual approach to map the relationship between the output and the original input of image and analyze the feature learning mechanism of DCNN in remote sensing scene recognition. The experimental results show that the recognition of complex remote sensing scene involves the process of learning the distribution of multiple objects. The joint distribution of multiple targets supports scene understanding.

5.1 CAM Visualized Analysis

In this paper, we use CAM (Zhou, Khosla et al. 2015) to visualize objects, which highly contributes to recognition. This method builds the mapping relationship between the feature and probability of the image output in the feature transfer process through the hidden layers and reflects the influence of the object on a heat map (Figure 8). The most important part in it is building class activation maps:

$$M_c = \sum_k w_k^c f_k(x, y) \tag{1}$$

where $M_c$ is the contribution of spatial position $(x, y)$ to the model predicted class c, $w_k^c$ is the weight of the kth neuron of the last convolution layer corresponding to class c, and $f_k(x, y)$ is the activation of the kth neuron of the last convolution layer at position $(x, y)$.

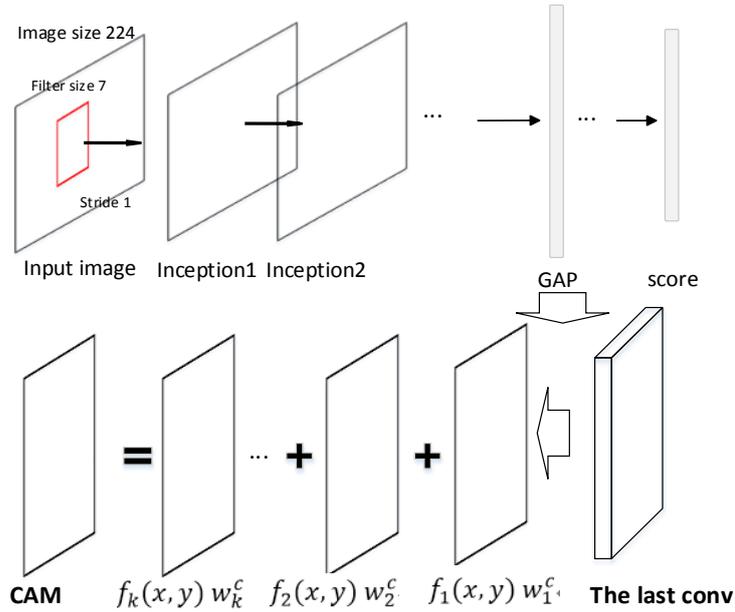

Figure 8. GoogleNet* and CAM calculation

5.2 Experimental analysis and conclusion

We train the model on the AID-22 Scene dataset using the GoogleNet* proposed in CAM. The network removed the auxiliary loss function based on GoogleNet and replaced the full connection layer with global average pooling (GAP). We test 72 samples of 22 classes and the overall accuracy achieves over 90%, which ensures that the model captures useful features of the sample. Table 4 shows the overall accuracy of the model.

Table 5. Accuracy of GoogleNet* tested on AID-22 Scene

| Model | Train accuracy | Test accuracy |
| --- | --- | --- |
| GoogleNet* | 94.15% | 91.82% |

Based on this model, we use the CAM to visualize the feature distribution of the 22 scene categories (Figure 8). We find that the DCNN-based scene recognition actually learns the object and the distribution among them as reflected in the spatial location by capturing the characteristics of single or multiple key objects in the scene. For example, in Figure 9, the area of the car is highlighted for the scene whose ground truth is parking, and this finding indicates that the model focuses on learning the characteristics of the car and the overall distributed characteristics of the cars in general. Similarly, the recognition of port scenarios is determined by waters and its surrounding areas. Figure 10 shows the CAM results of scenes with different complexities.

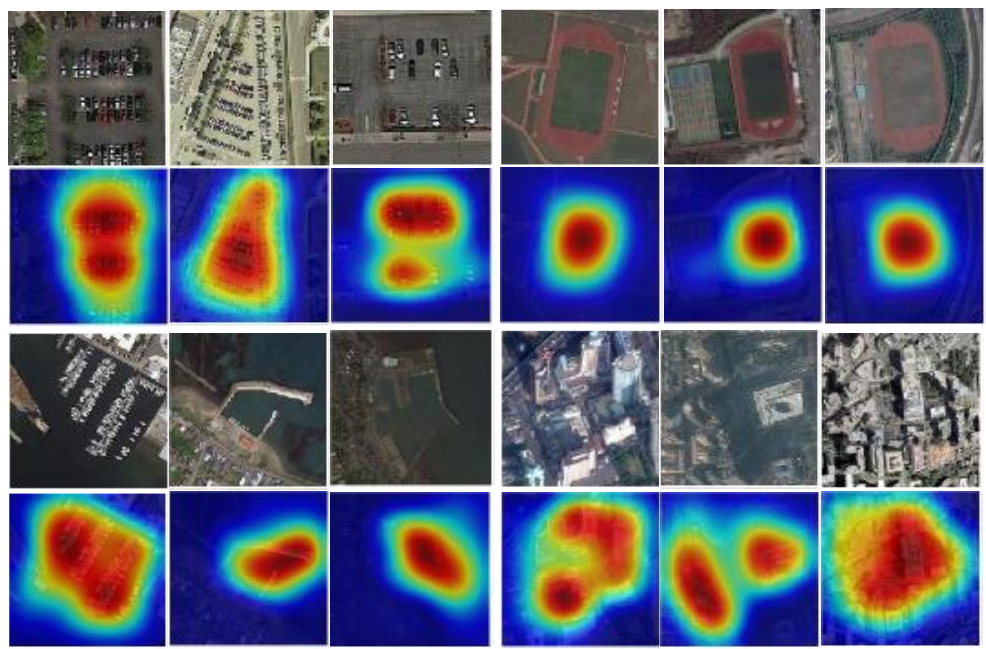

Figure 9. CAMs of four classes in the AID-22 scene. The warmer the color is, the greater the contribution of the region to the classification results. The lower the tone, the smaller the contribution of the region to the classification

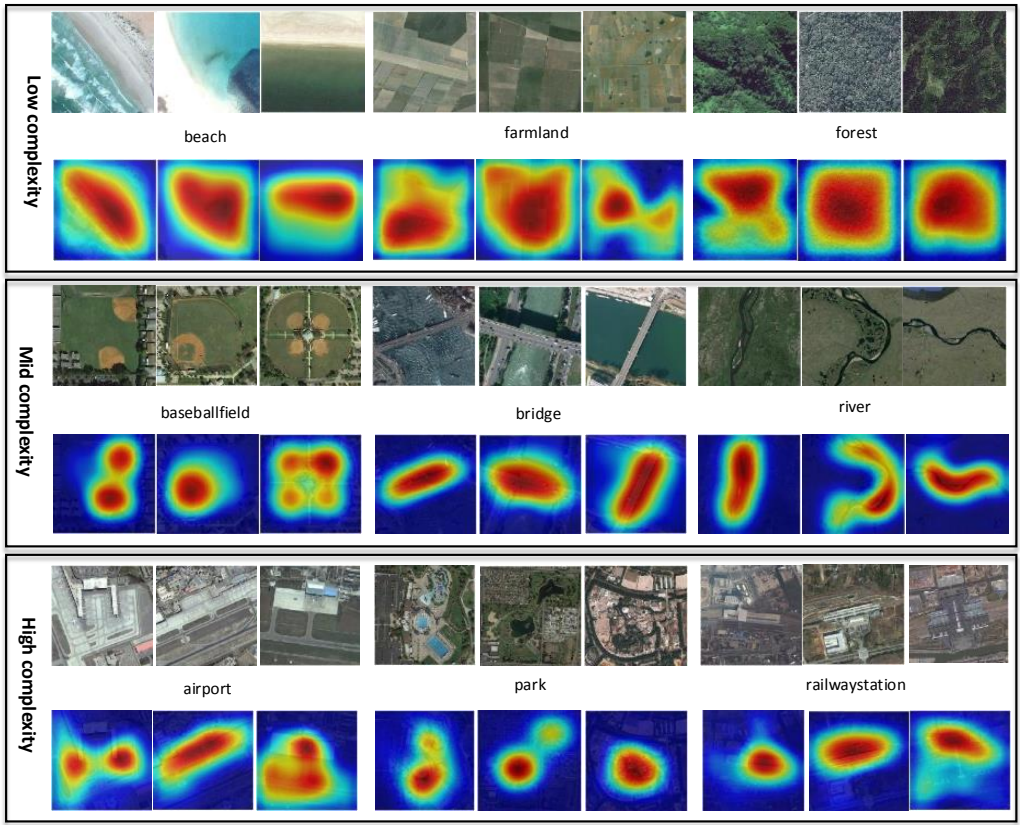

Figure 10. CAMs of different complexity scenarios. The top is the results of low complexity scene visualization; the middle is the moderate complex scene; the bottom is the highly complex scene

We analyze the relationship between the corresponding features and different output probabilities, and we find a strong correlation between the recognition result and the responsive mode of CAM. The model captures the characteristics of the corresponding target in spatial region, and the difference in spatial position is reflected in a responsive pattern, which ultimately affects scene recognition. Figure 11 shows the top-1 to top-5 output probabilities corresponding to the CAMs of a category, and the coast, sand, and waves all respond to the top-1; this finding indicates that the model abstracts the concept of the beach by learning the characteristics of the three categories of objects. The seawater and seawater edge have more intense responses in top-2 to demonstrate that the model mainly learns both types of objects, resulting in the identification of the sample as port category.

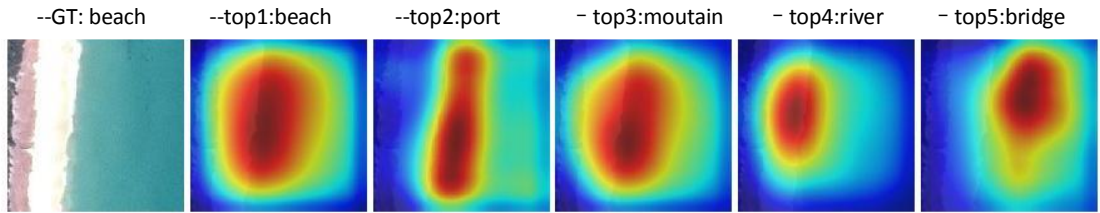

Figure 11. Top-5 CAMs of a sample (GT is beach)

To exclude this phenomenon only for a specific category, we randomly select 20 samples from per category to obtain the CAMs and statistics of the proportion of samples that respond to multiple targets in each scene (Figure 12). CAMs of 80% of the samples in the 15 categories, of which the total number is 22, responds to multiple objects. Meanwhile, in some scenes, the proportion of the samples which meet the pattern is lower than 50%. This result is mainly because these scenes express semantic information through single object, such as the center scene, which represents specific individual object (building).

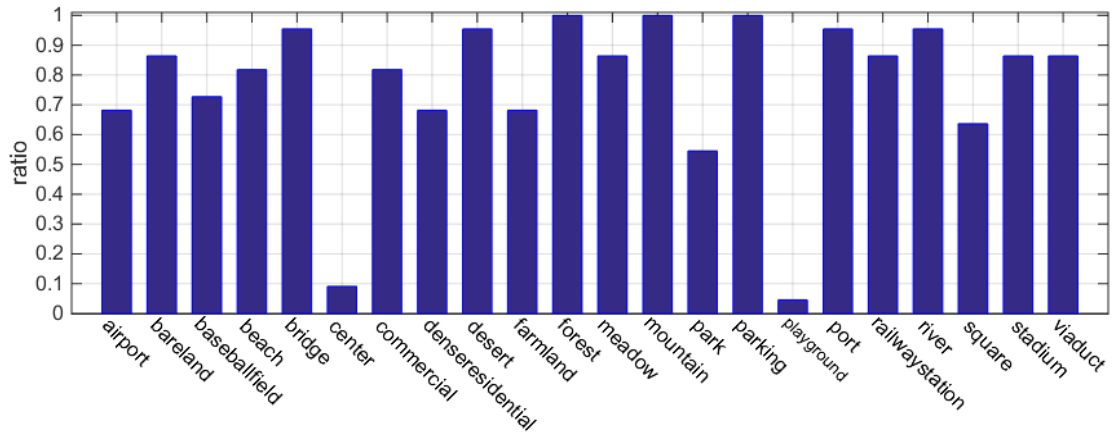

Figure 12: Percentage of CAMs that conform joint distribution. The horizontal axis represents the scene type, and the vertical axis represents the proportion of samples that responds to multiple objects in the CAMs in each category

We conduct similar experiments on the CID256 dataset, mask multiple objects of scene and analyze the relationship between the output probability of the model and responsive pattern of the CAMs (Figure 13). The result shows that the recognition of remote sensing scene has learned the comprehensive features of multiple targets and that a single target is insufficient to semantically represent the scene.

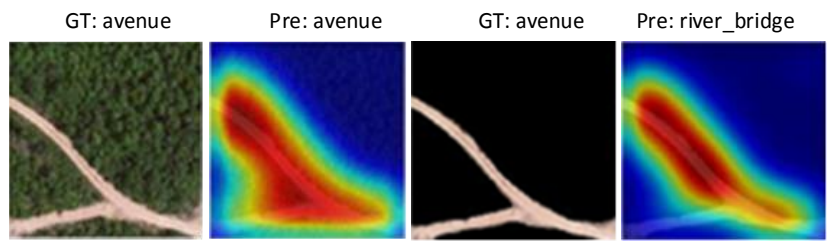

Figure 13: Occluding object of scene. Left: GT is the avenue that the model correctly identifies,

and CAM responds in the forest and road areas; right: block the area of forest, model incorrectly identifies it as river_bridge, and the corresponding CAM response area is mainly concentrated at the road partition.

## 6. Discussion and Future Work

Our work is inspired by the temporal and spatial imbalance of the attention placed in understanding the scene and focuses on exploring the depth and scale of the model in the task of remote sensing scene recognition from the perspective of scene complexity. A visualization method is used for quantitative and qualitative analyses of the mechanism in remote sensing scene recognition. The main conclusions are as follows:

1. Remote sensing scene recognition with different complexities depends on different network depths and sale characteristics. This means complex scene concept is the representation of multiple scale and level feature, there is implicit correlation between complexity and feature.
2. Complex scenes rely on the feature representation of multiple scales and the joint semantic support of multiple targets. We demonstrate that multiple scale involved model outperforms any other single network; and experiments based on CAMs indicates that the distribution of combing targets play significant role in learning mechanism.

In our next work, we will design an effective indicator to evaluate remote sensing scene complexity, and then design a network with an adaptive depth, which considers multiple scales. We will further explore the manner in which the mechanism of multi-objective joint probability can be combined with the generation model to help us better underst and a scene.